\documentclass[letterpaper, 10 pt, conference]{ieeeconf}  

\IEEEoverridecommandlockouts                              

\usepackage{times}

\usepackage{multicol}
\usepackage[bookmarks=true]{hyperref}

\usepackage{amsthm}
\usepackage{amsmath} 
\usepackage{amssymb}  
\usepackage{xcolor}

\usepackage{verbatim}
\usepackage{algorithm}
\usepackage{algorithmicx}
\usepackage[noend]{algpseudocode}

\usepackage{graphicx}
\usepackage{textcomp}
\usepackage{xcolor}
\usepackage{booktabs}
\usepackage{siunitx}

\usepackage{multirow,array}

\usepackage{subcaption}
\usepackage{graphicx}
\usepackage{hyperref}
\usepackage{footmisc}

\usepackage[utf8]{inputenc}
\usepackage[english]{babel}

\theoremstyle{definition}

\renewcommand{\algorithmicrequire}{\textbf{Input: }}
\renewcommand{\algorithmicensure}{\textbf{Output: }}

\title{\LARGE \bf
Contact Mode Guided Motion Planning for \\Quasidynamic Dexterous Manipulation in 3D
}

\author{Xianyi Cheng, Eric Huang, Yifan Hou, and Matthew T. Mason
\thanks{*This work was supported under NSF Grant IIS-1909021.}
\thanks{The authors are with Carnegie Mellon University, Pittsburgh, PA, 15213, USA. 
{\tt\small <xianyic, erich1, yifanh>@andrew.cmu.edu,  mattmason@cmu.edu}}%
}

\begin{document}
\maketitle

\begin{abstract}
This paper presents Contact Mode Guided Manipulation Planning (CMGMP) for 3D quasistatic and quasidynamic rigid body motion planning in dexterous manipulation. The CMGMP algorithm generates hybrid motion plans including both continuous state transitions and discrete contact mode switches, without the need for pre-specified contact sequences or pre-designed motion primitives. 
The key idea is to use automatically enumerated contact modes of environment-object contacts to guide the tree expansions during the search. 
Contact modes automatically synthesize manipulation primitives, while the sampling-based planning framework sequences those primitives into a coherent plan. We test our algorithm on fourteen 3D manipulation tasks, and validate our models by executing some plans open-loop on a real robot-manipulator system\footnote{ \label{myfootnote} The video is available at  \url{https://youtu.be/JuLlliG3vGc}}. 

\end{abstract}


\section{Introduction}
\label{sec:intro}


Dexterous manipulation planning is challenging in many aspects. 
The first challenge is the exploitation of dexterity. 
The dexterity in manipulation can come from the dexterity of robot hands (intrinsic dexterity) and the exploitation of the environments (extrinsic dexterity) \cite{ChavanDafle2014extrinsic}. A general dexterous manipulation planner needs to use both intrinsic and extrinsic dexterity cleverly. 
The second challenge is the contact-rich nature of manipulation. Making and breaking contacts bring more complexity into the system by changing its kinematics and dynamics. This hybrid nature makes planning through contacts difficult. 
The CMGMP aims to move one step closer towards general dexterous manipulation planning by considering the dexterity from both the robot hand and the environment, and planning among these contacts. 
\begin{figure}[t]
    \centering
    \includegraphics[width=0.7\columnwidth]{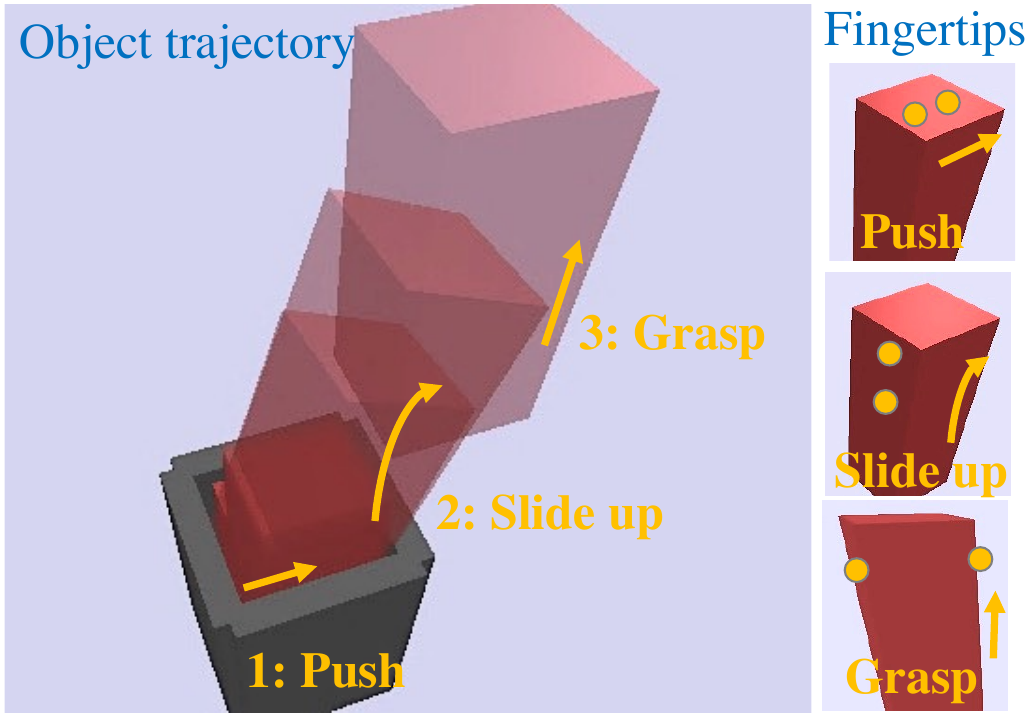}
    \caption{The goal is to get the object out of the box while the gaps are too small for both fingers to get in. Our method plans first to push the object on the top to create a wider gap that lets the fingers slide up the object from the side.
    }
    \label{fig:intro}
\end{figure}

The CMGMP algorithm is a Rapidly-Exploring Random Tree (RRT)~\cite{lavalle1998rapidly} based planner with automatically enumerated contact modes to guide tree extensions. The key idea is to use contact modes. Contact modes are helpful both in generating continuous motions and capturing discrete changes. Contact modes serve like automatically generated ``motion primitives'', guiding the planning of continuous motions in submanifolds of the configuration space. Compared with manually designed motion primitives, contact modes automatically generate more varied motions while requiring less engineering effort. In the discrete space, contact modes help find paths across manifolds. 
The set of kinematically feasible contact modes capture all possible discrete contact changes in the current system configuration because it enumerates all possible transitions of the contact states. Combined with sampling-based planning, contact modes boost the exploration over large continuous and discrete spaces for contact-rich motions. 

This paper presents the complete version of the CMGMP algorithm that can solve 3D quasistatic and quasidynamic dexterous manipulation tasks with a rigid object in a rigid environment, assuming non-sliding finger contacts, compatible with user-provided 3D object mesh models and robot-manipulator kinematic models. An example for peg-out-of-hole planning is shown in Figure \ref{fig:intro}. 
The planner generates the strategy of first pushing the object on the top to create a wider gap that lets the fingers slide up the object from the side, and then form a grasp. 
Compared to our 2D quasistatic version \cite{cheng2020contact}, this work demonstrates that this type of algorithms can solve a much more general range of manipulation tasks. 
To the best of our knowledge, the CMGMP is the first method capable of solving diverse dexterous manipulation tasks of such levels of complexity without any pre-designed skill or pre-specified modes. 
We believe the flexibility, simplicity, and general applicability demonstrated by this work bring us one step closer towards general dexterous robotic manipulation.


\section{Related Work}
\label{sec:literature}

\subsection{Manipulation Planning}
 

Efficient search and optimization algorithms have been developed to solve motion sequencing/planning problems~\cite{lozano2014sequential,toussaint2018differentiable}. Sampling-based planning methods like CBiRRT~\cite{berenson2009manipulation} and IMACS \cite{kingston2019exploring} explore the manifolds of known constraints. Most of these methods require predefined states or primitives. The solutions are also confined to be the combinatorics of predefined states/primitives. 
As existing taxonomies cannot even categorize all human grasping behaviors \cite{nakamura2017complexities}, it is therefore impractical to develop a skill library that fulfills the complexity and dexterity needed in general manipulation. 
In contrast, our work uses contact modes to automatically generate ``motion primitives'', allowing the planner to find a variety of lower-dimensional solutions.    

Contact formations~\cite{xiao2001automatic} have been explored in~\cite{trinkle1991dexterous,ji2001planning,tang2008automatic,lee2015hierarchical} to plan motions between two rigid bodies~\cite{tang2008automatic} and within a robot hand~\cite{trinkle1991dexterous, yashima2003randomized}. Contact information helps to decompose the search space into smaller chunks. Search and planning within contact formations are later combined into a complete solution. These methods all require to precompute all possible contact formations offline. 

Our previous work is a simplified version of the CMGMP \cite{cheng2020contact}. 
The addressed problems are limited to 2D domains. The quasistatic assumption prevents it from generating simple actions like dropping or toppling. 
This work presents a more complete version of the CMGMP by extending from 2D domains to 3D domains, from quasistatic to quasidynamic, from well-parameterized 2D object geometries to arbitray 3D object models, and from simple free-moving point manipulator model to more realistic user-defined robot kinematics and non-sliding finger contact models. 


\subsection{Contact-rich Motion Generation}
Trajectory optimization is effective in generating contact-rich motion plans. Contact-Invariant Optimization methods~\cite{mordatch2012contact,mordatch2012discovery} produce complex whole-body and manipulation behaviors in simulation, assuming soft contacts which may violate physics laws. Dynamic manipulation planning for rigid bodies is explored in \cite{posa2014direct,Sleiman2019CIO,doshi2020icra,aceituno-cabezas2020rss}, generating simple manipulation actions with small numbers of contact transitions, like pushing, pivoting, and grasping. Chen et al. \cite{chen2021trajectotree} interleaves tree search and trajectory optimization to efficiently solve a 2D in-hand manipulation problem. 
However, contact implicit trajectory optimization (CITO) methods \cite{posa2014direct} could be intractable without good initialization. 
To our best knowledge, there has not been any trajectory optimization method that can solve the tasks of similar level of complexities shown here.


\section{Problem Description}
\label{sec:problem_description}
\subsection{Inputs and Outputs}

The inputs to our method are the start and goal poses of the object, the geometries, and the properties of the object and the environment:
\begin{enumerate}
    \item \textbf{Object start pose}: 
    $q_{\mathrm{start}} \in SE(3)$.
    
    \item \textbf{Object goal region}: 
    $Q_{\mathrm{goal}} \subset SE(3)$.
    
    \item \textbf{Object properties}: a rigid body $\mathcal{O}$ with known geometry, mass distribution, and friction coefficients with environment $\mu_{\mathrm{env}}$ and with the manipulator $\mu_{\mathrm{mnp}}$.
    
    \item \textbf{Environment}: $\mathcal{E}$ with known geometries.
    
    \item \textbf{Manipulator model}: a user-defined model for the robot-manipulator system. The manipulator can make at most $N_{\mathrm{mnp}}$ contacts with the object.

\end{enumerate}

Our method outputs a trajectory $\pi$ that is a sequence of object motions, contacts, and contact modes. 
At step $t$, the trajectory $\pi(t)$ gives:  
\begin{enumerate}
    \item \textbf{Object motion}: object configuration $q(t)$ at step $t$. 
    \item \textbf{Environment contacts}: the contact points of the object with the environment $N_{\mathrm{env}}(t)$. The $k$th environment contact $c^{\mathrm{env}}_k(t)$ is specified by its contact location $p^{\mathrm{env}}_k(t)$ and contact normal $n^{\mathrm{env}}_k(t)$.
    
    \item \textbf{Manipulator configurations}: 
    the manipulator configuration $q^{\mathrm{mnp}}(t)$. The manipulator contacts with the object can be obtained from $q^{\mathrm{mnp}}(t)$ using forward kinematics: $c^{\mathrm{mnp}}(t) = [c^{\mathrm{mnp}}_1(t), c^{\mathrm{mnp}}_2(t), \dots, c^{\mathrm{mnp}}_{N_{\mathrm{mnp}}}(t)]$. 
    The $k$th manipulator contact is specified by its contact location $p^{\mathrm{mnp}}_k(t)$ and contact normal $n^{\mathrm{mnp}}_k(t)$.

    \item \textbf{Contact mode}: the 3D contact mode $m(t)$ of the environment contacts and the manipulator contacts. 
\end{enumerate}

\subsection{Assumptions}

In addition to the standard assumptions made within rigid body simulators (non-penetration, point contacts, polyhedral friction cones \cite{Lee2018, bullet}, etc.), we assume the following:
(1)     
The users can choose to enforce either quasistatic or quasidynamic assumptions. 
For quasistatic manipulation, inertial forces are negligible, and the object needs to be in force balance all the time. 
For quasidynamic manipulation, the tasks involve occasional brief dynamic periods. The accelerations do not integrate into significant velocities. Momentum and restitution of impact are negligible \cite{Mason}. 
In our experience, 
quasistatic \cite{Mason1986scope} and quasidynamic assumptions hold even
during fairly fast manipulator velocities. Moreover, manipulation motions synthesized in a quasistatic model are inherently safer.
(2) Environment-object contacts can have any modes, but
manipulator-object contacts are sticking only. We currently does not plan finger sliding motions.



\section{Contact Mode Based Manipulation Models}

\label{sec:model}

This section describes the force and motion models under contact modes with quasistatic and quasidynamic assumptions.

\subsection{Contact Modes in 3D}
\label{sec:3dmodes}

A contact mode describes the relative contact velocities for all the contacts in a system \cite{Mason}. 
A contact mode $m = [m_{\mathrm{cs}}, m_{\mathrm{ss}}]$ in 3D consists of two parts: the contacting/separating (CS) mode $m_{\mathrm{cs}}$ and the sticking/sliding (SS) mode $m_{\mathrm{ss}}$. 
The CS mode $m_{\mathrm{cs}}$ is the sign of the contact normal velocities $v_{c,n}$. For a system with $N$ contacts, we have $m_{\mathrm{cs}} = [\textit{sign}(v_{c,n}^i)] \in \{0,+\}^N$, where $v_{c,n}^i$ is the normal velocity for the $i$th contact in its own contact frame.
The SS mode $m_{\mathrm{ss}}$ of a CS mode identifies the directions of contact tangent velocities $v_{c,t}  \in \mathbf{R}^2$. 
The contact tangent planes are first being divided by $n_t$ equal angled hyperplanes $C_T = [C_T^1, \dots, C_T^{n_t}]^T$, we have $m_{\mathrm{ss}} = [\textit{sign}(C_T \cdot v_{c,t}^i)] \in \{\{-,0,+\}^{n_t}\}^N $. 


If we approximate unilateral contacts by linear complementarity constraints, each contact mode corresponds to a facet of the complementary cone \cite{berard2004contact}. Individually solving for each contact mode divides the complementarity constraint into easier sub-problems. 
Our previous work \cite{huang2020efficient} shows that the complexity of enumerating all 3D contact modes for one object is $\mathcal{O}(N^d)$, where $N$ is the number of contacts and $d$ is the effective degrees of freedom of the object. For example, $d$ is 6 for a free cube, 3 for a cube between two tight walls, 1 for a cube inside a tight square pipe. The number of contact modes for one object is usually 50 to 400, which makes 3D contact mode enumeration practical in our algorithm.


\subsection{3D Contact Mode Constraints on Velocities and Forces}
\label{sec:mode_constraits}
If we adopt the polyhedral approximation of friction cones \cite{stewart1996implicit}, given a contact mode $m \in \mathbf{M}$, we obtain linear constraints about the object motion and contact forces.

\noindent \textbf{Velocity Constraints}
Contact velocities for all the contacts in the contact frames can be written as:
\begin{equation}
\label{eqn:contactvel}
v_c = G^T v^o - \begin{bmatrix}J \dot{q} \\ 0\end{bmatrix}
\end{equation}
where $v^o$ is the object body velocity in the twist form; $G$ is the contact grasp map 
\cite{murray1994mathematical}; $\dot{q}$ is the manipulator joint velocity and $J$ is the manipulator's Jacobians; the $0$ part is for environment contacts since they are fixed. 

For the $i$th contact, its CS mode $m_{\mathrm{cs}}^i \in \{+, 0\}$ constrains the contact normal velocity $v_{c,n}^i$:
\begin{equation}
    \label{eqn:velocity_cs}
    \begin{cases}
    v_{c,n}^i > 0  \quad \text{if} \quad m_{\mathrm{cs}}^i = +\\
    v_{c,n}^i = 0  \quad \text{if} \quad m_{\mathrm{cs}}^i = 0
    \end{cases}
\end{equation}
The SS mode $m_{\mathrm{ss}}^i \in \{-, 0, +\}^{n_t} $ constrains the contact tangent velocity $ v_{c,t}^i$:
\begin{equation}
    \label{eqn:velocity_ss}
    \begin{cases}
    C_T^j \cdot v_{c,t}^i > 0  \quad \text{if} \quad m_{\mathrm{ss}}^{i,j} = +\\
    C_T^j \cdot v_{c,t}^i = 0  \quad \text{if} \quad m_{\mathrm{ss}}^{i,j} = 0\\
    C_T^j \cdot v_{c,t}^i < 0  \quad \text{if} \quad m_{\mathrm{ss}}^{i,j} = -
    \end{cases}
\end{equation}
where $C_T$ contains $n_t$ vectors that partition the contact tangent plane. If $n_t = 2$ (for a 4-sided polyhedral friction cone), we have $C_T^1 = [1,0]$ and $C_T^2 = [0,1]$. 


\noindent \textbf{Force Constraints}

Let the magnitudes of contact force of the $i$th contact be $\lambda^i = [\lambda_{t_1}^i, \lambda_{t_2}^i,\lambda_n^i]$, for two contact tangent directions and the contact normal direction respectively.

The CS mode decides whether there exists contact forces:
\begin{equation}
    \label{eqn:force_cs}
    \begin{cases}
    \lambda^i = 0   \quad \text{if} \quad m_{\mathrm{cs}}^i = +\\
    \lambda_{n}^i > 0  \quad \text{if} \quad m_{\mathrm{cs}}^i = 0
    \end{cases}
\end{equation}

If $m_{\mathrm{cs}}^i$ is $0$, there exists contact forces.
There are two different conditions: the contact is sticking ($\forall j, m_{\mathrm{ss}}^{i,j} = 0$), and sliding ($\exists j, m_{\mathrm{ss}}^{i,j} \neq 0$). 

When the contact is sticking, $\lambda_{t_1}^i$ and $\lambda_{t_2}^i$ are the contact tangent forces in $x$ and $y$ axis. The contact force should be in the polyhedral friction cone of the friction coefficient $\mu$: 

\begin{equation}
    \label{eqn:force_ss_sticking}
    \begin{bmatrix} C_T & \mu \\ -C_T & \mu \end{bmatrix} \cdot \lambda^i > 0
\end{equation}

If the contact is sliding, due to the maximum dissipation law, the tangent force should be in the opposite direction of the sliding velocity. 
Let $\{h^{i,j}\}$ to be the edge(s) of the 1D/2D contact tangent velocity cone of $m_{\mathrm{ss}}^i$, $\lambda_{t_1}^i$ and(or) $\lambda_{t_2}^i$ are contact force magnitudes in the $\{h^{i,j}\}$ direction(s). In our polyhedral approximation, the contact tangent force $f^i_t$ should be in the opposite cone of the contact sliding velocity cone: 
\begin{equation}
\label{eqn:sliding_force}
f^i_t = -\sum_j \lambda_{t_j}^i h^{i,j}, \lambda_{t_j}^i > 0    
\end{equation}
where the edge(s) $\{h^{i,j}\}$ for all $m_{\mathrm{ss}}^i$ can be precomputed by $C_T$. 
From the Coulomb friction law, we have:  
\begin{equation}
    \label{eqn:force_ss_sliding}
    \begin{split}
    \lambda_{t_j}^i > 0, \forall j \\
    \mu \lambda_{n}^i - \sum_{j=1}^K \lambda_{t_j}^i = 0
    \end{split}
\end{equation}

Putting the velocity and force constraints for all contacts together (Equation \ref{eqn:contactvel},\ref{eqn:velocity_cs},\ref{eqn:velocity_ss},\ref{eqn:force_cs},\ref{eqn:force_ss_sticking},\ref{eqn:force_ss_sliding}), letting $\lambda$ be the magnitudes of all contact force directions, we get a set of linear equations and inequalities from contact mode constraints: 


\begin{equation}
    \label{eqn:together}
    A_{\mathrm{ineq}} \begin{bmatrix}v & \lambda \end{bmatrix}^T > b_{\mathrm{ineq}}, 
    A_{\mathrm{eq}} \begin{bmatrix}v & \lambda \end{bmatrix}^T = b_{\mathrm{eq}}
\end{equation}


\subsection{Quasistatic Assumption}
\label{sec:quasistatic}
In quasistatic manipulation, the object should always be in a force balance. 
The force balance equation is written as:
\begin{equation}
\label{eqn:static}
   \begin{bmatrix}G_1 h_1, G_2 h_2, \dots\end{bmatrix} \cdot \begin{bmatrix}\lambda_1, \lambda_2, \dots \end{bmatrix}^T + F_{\mathrm{external}} = 0
\end{equation}
where $\begin{bmatrix}\lambda_1, \lambda_2, \dots \end{bmatrix}^T$ are the magnitudes of forces along active contact force directions $\begin{bmatrix}h_1, h_2, \dots \end{bmatrix}^T$ determined by contact modes as described in Section \ref{sec:mode_constraits}. $\begin{bmatrix}G_1, G_2, \dots \end{bmatrix}^T$ are the contact grasp maps.
$F_{\mathrm{external}}$ includes other forces on the object, such as gravity and other applied forces.

\subsection{Quasidynamic Assumption} 

Quasidynamic assumption relaxes the requirement for object being in force balance, allowing short periods of dynamic motions. We assume accelerations do not integrate into significant velocities. In numerical integration, the object velocity from the previous timestep is $0$. The equations of motions become: 

\begin{equation}
\label{eqn:quasidynamic}
   M_o \dot{v^o} = \begin{bmatrix}G_1 h_1, G_2 h_2, \dots\end{bmatrix} \cdot \begin{bmatrix}\lambda_1, \lambda_2, \dots \end{bmatrix}^T + F_{\mathrm{external}}
\end{equation}
In discrete time, the object acceleration $\dot{v^o}$ can be written as $\frac{v^o}{h}$, where $h$ is the step size.

\subsection{Solve for Desired Motions at Every Timestep}
In Section~\ref{sec:forwardintegrate}, we use numerical integration to compute the object trajectories. Here we describe how we find the object motion at each timestep using the force and motion models derived above. 
We solve a quadratic program to obtain the optimal motion given a desired velocity $v^o_{\mathrm{des}}$:
\begin{equation}
\label{eqn:optvel}
\begin{array}{ll}
     & \min\limits_{v^o, \dot{q}, \lambda,} \| v^o_{\mathrm{des}} - v^o \|_2^2 + \epsilon \lambda^T \lambda \\
    \text{s.t.} & \small \text{Equation \ref{eqn:together} (contact mode constraints)} \\
    \small &\small \text{Equation \ref{eqn:static} (quasistatic)}/\text{Equation \ref{eqn:quasidynamic} (quasidynamic)} 
\end{array}
\end{equation}
where $\epsilon \lambda^T \lambda$ is a regularization term on the contact forces. 


\section{The CMGMP Algorithm}
\label{sec:planner}


\subsection{Planning Framework Overview}
\label{sec:framework}

Algorithm \ref{alg:rrt} presents our planning framework. The function \textsc{sample-object-config} samples an object configuration $q_{\mathrm{rand}}$, with a user-defined possibility $p$ of being a random sample and $1-p$ of being $q_{\mathrm{goal}}$. For $q_{\mathrm{rand}}$, its nearest neighbor $q_{\mathrm{near}}$ in the tree $\mathcal{T}$ is found through a weighted $SE(3)$ metric: 
\begin{equation}
\label{eqn:se3metric}
    \mathrm{d}(q_1, q_2) = \mathrm{d}_{\mathrm{trans}}(q_1,q_2) + w_r \cdot \mathrm{d}_{\mathrm{angle}}(q_1,q_2)
\end{equation}
where $w_r$ is the weight that indicates the importance of rotation in the system, and $d_{\mathrm{trans}}$ is the translation and $d_{\mathrm{angle}}$ measures the rotation angle between two configurations. Collision checking is performed to obtain environment contacts $c^{\mathrm{env}}$. 
The function $\textsc{cs-mode-enumeration}$ enumerates all feasible CS modes $\mathbf{M}_{\mathrm{cs}}$ for $c^{\mathrm{env}}$. 
Under each CS mode $m_{\mathrm{cs}} \in \mathbf{M}_{\mathrm{cs}}$, the function \textsc{extend} expands the tree from $q_{\mathrm{near}}$ towards $q_{\mathrm{rand}}$ for a user-defined maximum distance. 

The \textsc{extend} function has three major steps: (1) \textsc{best-ss-mode}: chooses a best SS mode given the desired object motion; (2) \textsc{relocate-manipulator}: tries to relocate fingertip contacts when necessary; (3) \textsc{project-integrate}: generates motions that move towards $q_{\mathrm{rand}}$ under a contact mode. A successful extension adds a new node $q_{\mathrm{new}}$ and a new edge to $\mathcal{T}$. 
The following subsections include more details of these subcomponents in \textsc{extend}.
\begin{figure}[t]
\vspace{-0.25cm}
\begin{algorithm}[H]
    \small
    \caption{the CMGMP algorithm} \label{alg:rrt}
    \algorithmicrequire $q_{\mathrm{start}}$, $q_{\mathrm{goal}}$ \newline
    \algorithmicensure tree $\mathcal{T}$
    \begin{algorithmic}[1]
        \State $\mathcal{T}\text{.add-node}(q_{\mathrm{start}})$ 
        \While {(\textit{Time limit has not been reached})}
        \State $q_{\mathrm{rand}} \gets \Call{sample-object-config}{q_{\mathrm{goal}}}$
         
        \State $q_{\mathrm{near}}\gets \Call{nearest-neighbor}{\mathcal{T},q_{\mathrm{rand}}}$
        \State $c^{\mathrm{env}} \gets \Call{collision-detection}{q_{\mathrm{near}}}$
        \State $\mathbf{M}_{\mathrm{cs}} \gets \Call{cs-mode-enumeration}{c^{\mathrm{env}}}$
        \For {$m_{\mathrm{cs}} \in \mathbf{M}_{\mathrm{cs}}$} 
        \State {// Iterate all CS modes}
        \State $\Call{extend}{m_{\mathrm{cs}}, q_{\mathrm{near}}, q_{\mathrm{rand}}, c^{\mathrm{env}}}$ 
   
        \EndFor
        \EndWhile
        
        \State \Return $\mathcal{T}$
    
    \end{algorithmic}
\end{algorithm}
\vspace{-0.3cm}
\end{figure}

\begin{figure}[t]
\vspace{-0.25cm}
\begin{algorithm}[H]
    \small
    \vspace{0.2cm}
    \caption{\textsc{extend} function} \label{alg:extend}
    \algorithmicrequire $m_{\mathrm{cs}}$, $q_{\mathrm{near}}$, $q_{\mathrm{rand}}$, $c^{\mathrm{mnp}}$, $c^{\mathrm{env}}$
    \begin{algorithmic}[1]
    \State $m_{\mathrm{ss}} \gets \Call{best-ss-mode}{q_{\mathrm{near}}, q_{\mathrm{rand}}, c^{\mathrm{mnp}}, c^{\mathrm{env}}}$
    \State $m \gets \Call{full-mode}{[m_{\mathrm{cs}}, m_{\mathrm{ss}}]}$
    \State $q^{\mathrm{mnp}} \gets \Call{previous-manipulator-config}{q_{\mathrm{near}}}$
    \If{not $\Call{motion-feasible}{m, q_{\mathrm{near}}, q^{\mathrm{mnp}}, c^{\mathrm{env}}}$ ...\\ or $\Call{manipulator-feasible}{q^{\mathrm{mnp}}}$}
    \State $q^{\mathrm{mnp}}_{\mathrm{new}} \gets \Call{relocate-manipulator}{q_{\mathrm{near}}, q^{\mathrm{mnp}}, m}$
    \Else 
    \State $q^{\mathrm{mnp}}_{\mathrm{new}} \gets q^{\mathrm{mnp}}$
    \EndIf
    \State $q_{\mathrm{new}} \gets \Call{project-integrate}{q_{\mathrm{near}}, q_{\mathrm{rand}}, q^{\mathrm{mnp}}_{\mathrm{new}}, m}$
     \If {$q_{\mathrm{new}} \neq q_{\mathrm{near}}$} 
    \State $\Call{assign-finger-contact-to-node}{q_{\mathrm{new}},q^{\mathrm{mnp}}_{\mathrm{new}}}$ 
    \State $\mathcal{T}\text{.add-node}(q_{\mathrm{new}})$
    \State $\mathcal{T}\text{.add-edge}(q_{\mathrm{near}},q_{\mathrm{new}})$
    \EndIf
    \State \Return 
    \end{algorithmic}
\end{algorithm}
\vspace{-0.3cm}
\end{figure}

\subsection{Extend for Every CS Mode and Filter SS Modes}
\label{sec:bestss}

A CS mode indicates a transition between two contact states. For every $q_{\mathrm{near}}$, we extend all its CS modes, so that the planner can explore all the next contact states from the current contact state. 
For each CS mode, we don't need to explore all its SS modes. The SS modes are approximations to the infinite number of sliding directions and polyhedral friction cones. We only need to select the SS mode in the best approximation cone for a desired object motion. In addition, to find solutions that lie in the lowest-dimensional space for a CS mode, such as an object pivoting around two sticking contacts, we also extend the all-sticking SS mode for each CS mode. 

We select the best SS mode that has the smallest cost. In the function \textsc{best-ss-mode}, for every SS mode, a quadratic programming is solved to get the closest velocity to the goal velocity under the contact mode velocity constraints (Equation \ref{eqn:optvel}). In practice, for the cost function in Equation \ref{eqn:optvel}, we assign a weight for angular velocity, often the same as $w_a$ in the SE(3) metric in Equation \ref{eqn:se3metric}, so that the cost function becomes: 
\begin{equation}
\label{eqn:weightedcost}
cost = \| v^o_{\mathrm{des}(v)} - v^{o(v)} \|_2^2 + w_a \| v^o_{\mathrm{des}(\omega)} - v^{o(\omega)} \|_2^2    
\end{equation}
where $v^{o(v)}$ is the translational velocity and $v^{o(\omega)}$ is the angular velocity. 

\subsection{Projected Integration}
\label{sec:forwardintegrate}

Starting from an object configuration $q_{\mathrm{near}}$, all reachable object configurations under the constraints by a contact mode $m$ form a manifold with boundary in the object configuration space $\mathcal{M}_m$. 
This projected integration process finds the furthest configuration $q_{\mathrm{new}}$ that the object could reach from $q_{\mathrm{near}}$ towards $q_{\mathrm{rand}}$ on $\mathcal{M}_m$.

At time-step $k$, we first update the desired object velocity $v^o_{\mathrm{des}}$ as the body velocity between $q_k$ and $q_{\mathrm{rand}}$ in the twist form. 
Then, Equation \ref{eqn:optvel} obtains $v^o_k$ by substantially projecting $v^o_{\mathrm{des}}$ onto the tangent space at $q_k$ of $\mathcal{M}_m$. We integrate $v^o_k$ by the first order forward Euler method:
\begin{equation}
    q_{k+1} = Tr(q_k, h v^o_k)
\end{equation}
where $h$ is the size of the time-steps, $Tr$ is the rigid body transformation computed from the body velocity $h v^o_k$, applied on $q_k$ \cite{murray1994mathematical}. The environment contacts in the constraints of Equation \ref{eqn:optvel} are updated every time-step by collision checking.

The projected integration stops when:
(1) $v^o_{\mathrm{des}}$ is zero: $q_k$ is the closest configuration on $\mathcal{M}_m$ to $q_{\mathrm{rand}}$. 
(2) No feasible $v^o_k$: moving towards $q_{\mathrm{rand}}$ is not feasible with current contact mode under the quasistatic or quasidynamic assumption. 
(3) The robot collides with the environment, or new object-environment contacts are made. 
(4) No solution exists for the robot inverse kinematics (IK). 
(5) User-defined maximum translation or rotation for a step has been reached.

To address the constraint drift problem, we project the object configuration back to the contact manifold every several time-steps through a correction velocity $v_{\mathrm{cor}}$: 
\begin{equation}
\label{eqn:velocitycorr}
    \min\limits_{v_{\mathrm{cor}}} \| G_N v_{\mathrm{cor}} - d^c \| + \epsilon \| v_{\mathrm{cor}} \|
\end{equation}
we have $G_N = \begin{bmatrix} C_N G_1^T,  \dots, C_N G_i^T, \dots \end{bmatrix}^T$ where $C_N = [0,0,1,0,0,0]$ and $G_i$ is the grasp matrix of the $i$th contact. $d^c = [d_1^c, \dots d_i^c]^T$ are the contact distances. The solution is $v_{\mathrm{cor}} = (G_N^T G_N + \epsilon I)^{-1} G_N^T d^c$.

\subsection{Planning Finger Locations}
\label{sec:fingersampling}

The function \textsc{relocate-manipulator} explores new finger contacts and robot configurations when: (1) the initial robot configuration is not assigned; (2) the current robot configuration will be in a collision or out of workspace; (3) current finger contacts no longer provides desired motions; (4) being randomly sampled to. 

This function is based on rejection sampling. We pre-compute a set of evenly distributed 
finger placements based on the object mesh model using supervoxel clustering \cite{papon2013voxel}. In practice, we store about 200 finger placements per object. 
During the planning, we first randomly select some fingers to relocate to new contact locations on the object. The new contacts should provide the desired motion under the current contact mode. The transition is feasible if the remaining finger contacts and new contacts from previously unlocated fingers can keep the object in force balance (quasistatic and quasidynamic) or can still provide the desired motion (quasidynamic). Next, we solve the robot IK. If there is a reachable collision-free robot configuration (a robot-specific finger relocation sub-planner is required), the new sample is accepted. Otherwise, the algorithm repeats this process until it gets a feasible sample or reaches the iteration limit.

\section{Results}
\label{sec:results}

\begin{table*}[h]
    \centering
    \vspace{0.2cm}
    \begin{tabular}{lllll}
    \toprule
    \# & Task Name & Assumption & Robot & Description \\ \midrule
    1 & Peg-out-of-hole & Quasistatic & 3 free-moving balls & The small gaps prevent a direct pickup\\
    2 & Bookshelf & Quasistatic & \vtop{\hbox{\strut 3 free-moving ball }\hbox{\strut a parallel jaw gripper }\hbox{\strut the DDHand}} & Take a book from a bookshelf\\
    3 & Pick up a card & Quasistatic & \vtop{\hbox{\strut 3 free-moving ball }\hbox{\strut the DDHand}} & The object is too thin or too wide to grasp\\
    4 & Grab a bottle & Quasistatic & 3 free-moving balls & The bottle is surrounded by other bottles \\
    5-7 & Object Reorientation & Quasidynamic & 2 free-moving balls & Cube (Task 5), T-block (Task 6), Hexbolt (Task 7) \\
    8 & Rolling & Quasistatic & 2 free-moving balls & Reorient an object with a smooth surface\\
    9 & Cube-and-Wall & Quasistatic & One free-moving ball &  The object is too heavy to pick up\\
    10 & Cube-and-Stairs & Quasistatic & 2 free-moving ball & The object is too heavy to pick up\\
    11 & Regrasp a Cube & Quasidynamic & A parallel jaw gripper & Use a parallel jaw gripper to reorient an object\\
    12 & Placedown & Quasidynamic & 2 free-moving balls &  Place down a thin object in a grasp\\
    13 & Flip-and-Pinch & Quasidynamic & \vtop{\hbox{\strut 2 free-moving ball }\hbox{\strut the DDHand}} & Flip then pinch a thin object on the table with two fingers\\
    14 & Object Reorientation & Quadidynamic & A robot arm + a rod finger & Reorient a cube considering robot kinematics and collision\\

    \bottomrule
    \end{tabular}
    \caption{Brief description of some selected test tasks.}
    \label{tab:tasks}
\end{table*}
\begin{table*}[th]
    \centering
    \resizebox{\textwidth}{!}{
    \begin{tabular}{c|cccccccccccccc}
    \toprule
    Task & 1 & 2 & 3 & 4 & 5 & 6 & 7 & 8 & 9 & 10 & 11 & 12 & 13 & 14 \\  \midrule

    Success  & 10/10 & 10/10 & 10/10 & 10/10 & 30/30 & 27/30 & 29/30 & 10/10 & 9/10 & 6/10  & 30/30 & 10/10 & 10/10 & 10/10\\ 
    
    {Time (second)}  & 5.4$\pm$1.1 & 2.7$\pm$1.3 & 3.9$\pm$1.0 & 1.7$\pm$0.8  & 4.8$\pm$2.8 & 11$\pm$10 & 2.8$\pm$1.7 & 0.6$\pm$0.3 & 8.3$\pm$3.9 & 46$\pm$11 & 14$\pm$9.0 & 12$\pm$7.0 & 9.0$\pm$2.8 & 23$\pm$14\\

    Nodes in Solution   & 9$\pm$2 & 7$\pm$1 & 8$\pm$2 & 6$\pm$1 & 7$\pm$3 & 11$\pm$6 & 15$\pm$8 & 7$\pm$3 & 12$\pm$8 & 42$\pm$12  & 6$\pm$2  & 12$\pm$4 & 12$\pm$5 & 4$\pm$0\\
    
    Nodes in Tree & 55$\pm$16 & 28$\pm$10 & 38$\pm$16 & 27$\pm$20 & 52$\pm$23 & 78$\pm$62 & 77$\pm$38 & 18$\pm$8 & 82$\pm$65 & 437$\pm$95 & 41$\pm$20 & 68$\pm$27 & 75$\pm$26 & 34$\pm$19 \\
    \bottomrule
    \end{tabular}}
    \caption{Planning statistics. A run is successful if a solution can be found within 100 iterations (200 for Task 10). ``Time'', ``Nodes in Solution'' and ``Nodes in Tree'' are in the format of ``mean''$\pm$``standard deviation'' for successful runs.}
    \label{tab:experiment}
\end{table*}

\subsection{Planning Results for Simulated Manipulation Tasks}

\begin{figure}[t]
    \centering
    \vspace{0.2cm}
    \includegraphics[width=0.8\columnwidth]{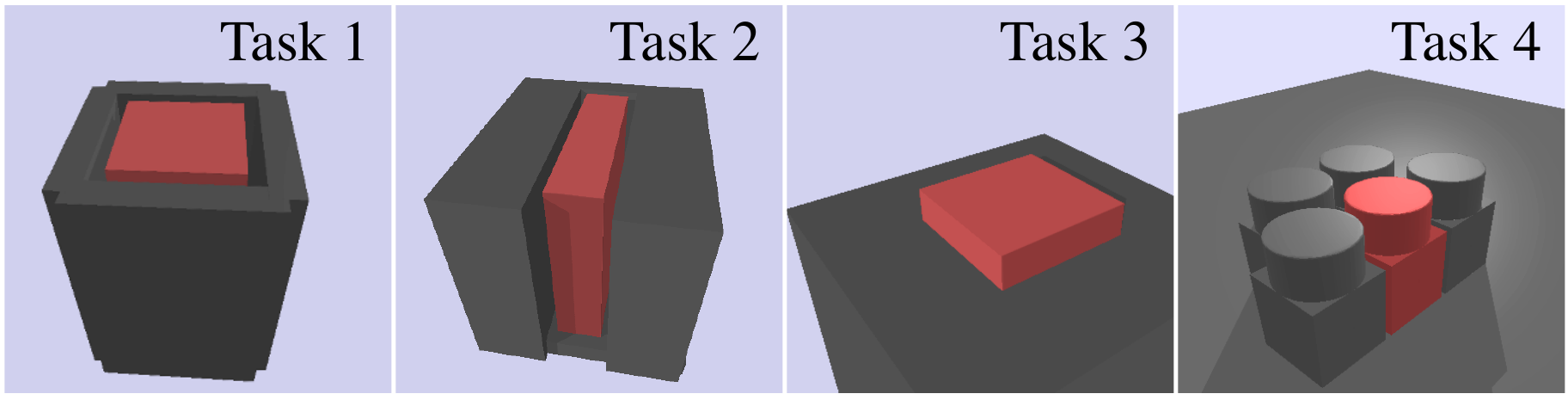}
    \includegraphics[width=0.8\columnwidth]{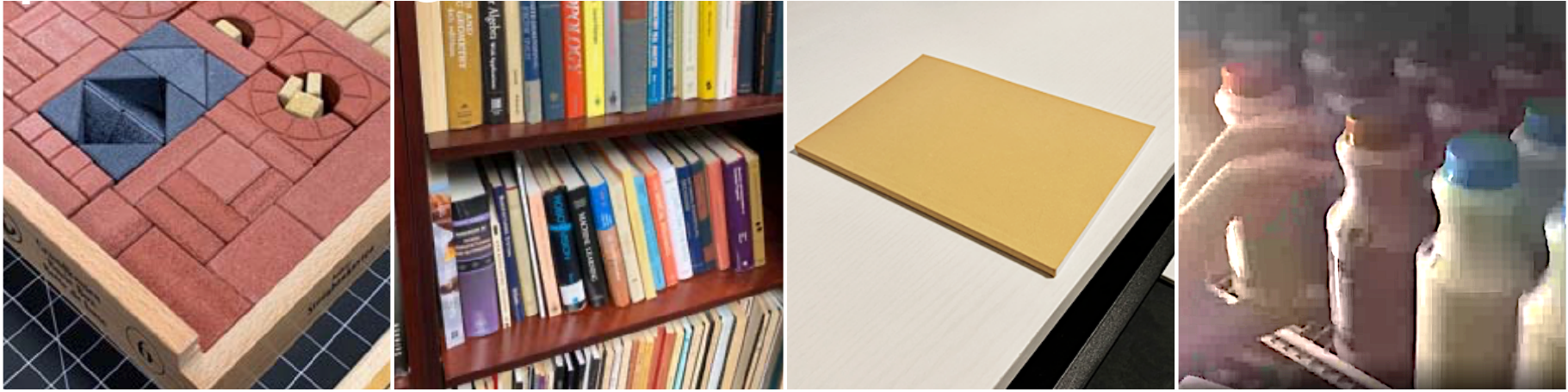}
    \caption{Task 1-4. The lower row shows the corresponding real-life scenarios. Task 1: get the first block out of packed blocks. Task 2: take a book from a bookshelf. Task 3: pick up a very thin book from the table. Task 4: grab a bottle from the fridge full of bottles (the photo is from \cite{nakamura2017complexities}). }
    \label{fig:task1-4}

\end{figure}

\begin{figure}[t]
    \centering
    \includegraphics[width=0.9\columnwidth]{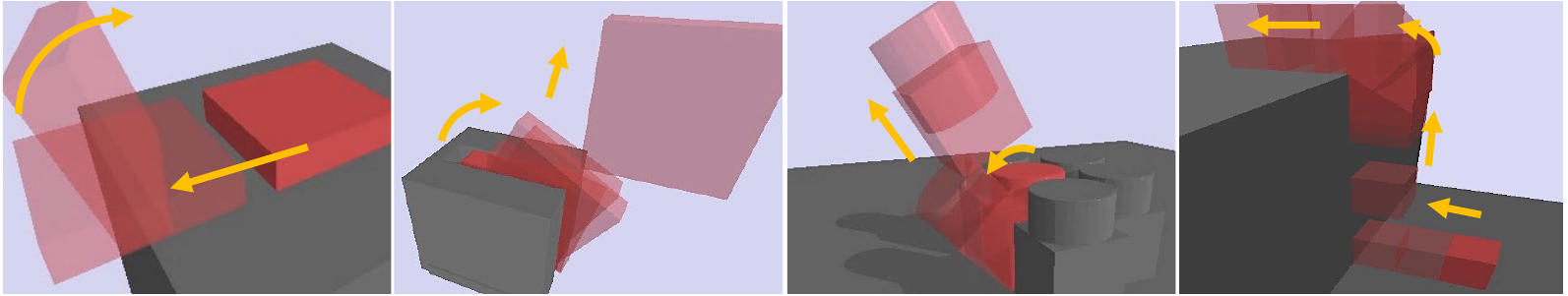}
    \caption{Planned object trajectories for Task 2, 3, 4 and 8.}
    \label{fig:traj}
\end{figure}

We test our planner on manipulation tasks that vary in object shapes, environments, model assumptions, manipulator types, etc. 
All the tasks need dexterous maneuvers due to specific task constraints. 
Table~\ref{tab:tasks} gives brief descriptions on 14 representative simulated tasks. Figure~\ref{fig:task1-4} shows the real-life scenarios for Task 1-4. 
The supplementary video\footref{myfootnote} and Figure~\ref{fig:traj} visualize some planning results. 

\begin{figure}[t]
    \centering
    \includegraphics[width=0.9\columnwidth]{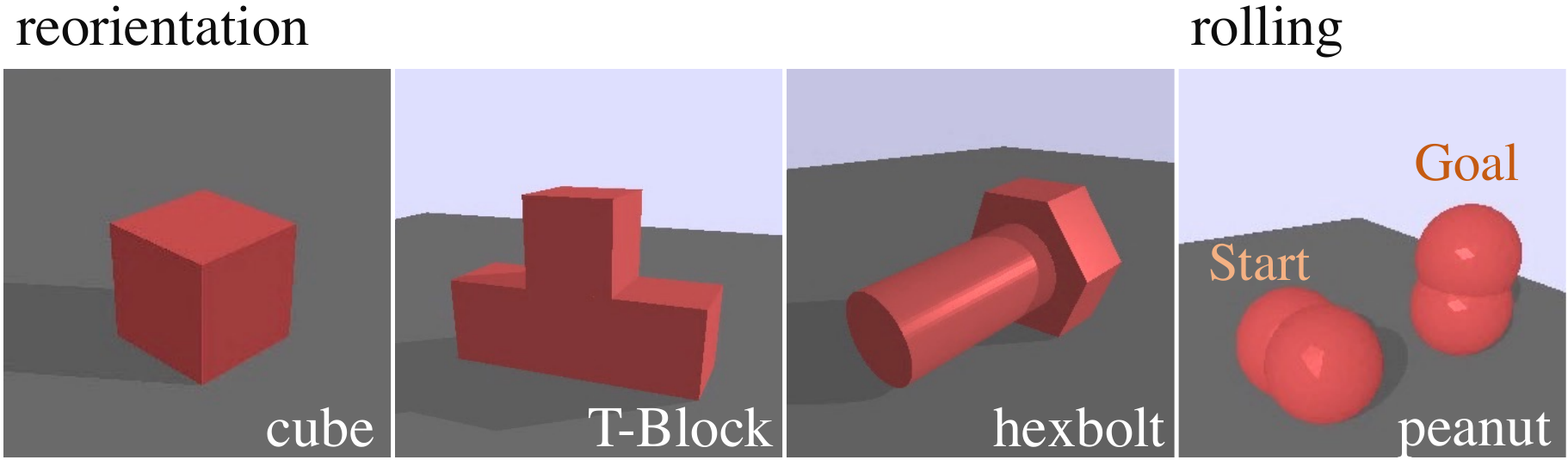}
    \caption{Test objects for object reorientation (Task 5-7) and rolling (Task 8).}
    \label{fig:objects}

\end{figure}

Task 1 (peg-out-of-hole), as explained in Section~\ref{sec:intro}, is commonly encountered as unpacking in our daily life. 
Task 2 (bookshelf) requires the book to be first slid/pulled out to form a grasp. We filmed a human participant taking a book with all the fingers and with only two fingers. The strategies by our planner are surprisingly similar to human strategies. 
In Task 3 (pick up a blade), 
our planner consistently generates a strategy of sliding the blade to the edge of the table to expose its bottom for grasping, similar to the slide-to-edge grasp in \cite{eppner2015exploitation}.
Task 4 (take a bottle) is inspired by a human grasping behavior, ``simultaneous levering out and grasp formation'', observed in \cite{nakamura2017complexities} when a human took a bottle of drink from the fridge. Our planner finds similar strategies: the manipulator first pivots or pulls out the object to create more space to form a three-finger grasp. 
All the strategies generated by our planner for Task 1-4 leverage environmental contacts; they may be seen as constraint-exploiting grasps \cite{eppner2015exploitation} commonly observed in human grasping behaviors. 
Our planner also generates strategies to reorient and to roll objects of different shapes in Figure~\ref{fig:objects}. 
Task 8-9 demonstrate that this planner can plan over many contact changes, although in practice it should not be directly given such long horizon tasks. 
Task 11 - 14 exhibit the quasidynamic strategies that exploit the gravity of the objects. For example, the robot can move the fingers away and let the object drop. Another example is Task 13 inspired by the flip-and-pinch strategy from \cite{odhner2013open}.



\subsection{Implementation and Planning Statistics}

We implement the algorithm using C++ and Dart \cite{lee2018dart}, with Bullet \cite{bullet} for collision detection. The code is available at \url{https://github.com/xianyicheng/cmgmp}. Table \ref{tab:experiment} shows the planning statistics provided by a computer with the Intel Core i9-10900K 3.70GHz CPU. Our planner generates most contact-rich plans in several seconds. The number of nodes in the tree is small, indicating that the exploration is efficient. 

To set up new tasks, 
the users need to adjust some parameters to reasonable ranges according to the tasks (no need for careful tuning): (1) $w_a$ for the distance metrics used in Equation \ref{eqn:se3metric} and \ref{eqn:weightedcost}, which should be the product of the object characteristic length and the estimated importance of rotations. (2) the maximum translational and rotational distance for \textsc{extend} to move towards $q_{\mathrm{rand}}$.

\subsection{Real Robot Experiments}

\begin{figure}[t]
    \centering
    \includegraphics[width=0.7\columnwidth]{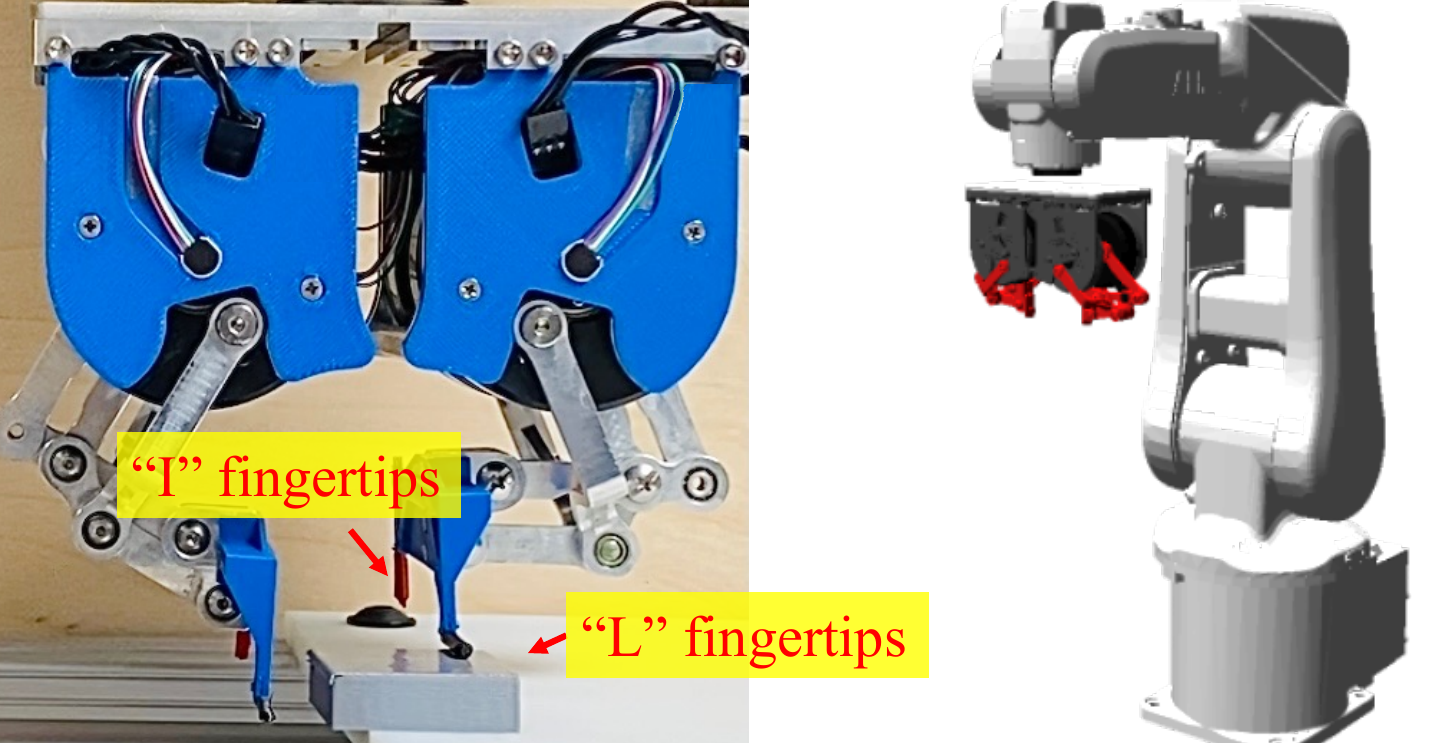}
    \caption{Left: The Dexterous DDHand executing Task 3. Right: the hand-robot system.}
    \label{fig:robot}

\end{figure}

To validate our models, we planned for Task 2 (bookshelf), Task 3 (pickup a blade), and Task 13 (flip-and-pinch) on a Dexterous Direct Drive Hand (DDHand) \cite{bhatia2019direct} mounted on an ABB IRB-120 robot arm, as shown in Figure~\ref{fig:robot}. The hand has two fingers, each with two degrees of freedom. There are two types of fingertips, the ``L'' type for Task 3 and 13, and the ``I'' type for Task 2. 
We provide the planner with the hand IK model and contact models for the fingertips. An ``L'' finger has a line contact model approximated by two point contacts. An ``I'' finger has a patch contact model, approximated by three point contacts. In the supplementary video, we run the planned robot trajectories only with robot position control. Although some runs are successful, the system is sensitive to uncertainties like object initial position errors. Robust executions will need controllers with force control \cite{hou2019robust} and force and vision feedbacks.

\section{Conclusion and Discussion}
\label{sec:conclusion}

We present the CMGMP framework that uses contact mode as guidance to generate quasistatic and quasidynamic dexterous manipulation motions in 3D. The strategies by our planner effectively leverage the environment as an external source for manipulation. 
For more general manipulation planning, we conclude several directions for future work: 
(1) Planning for fully dynamic manipulation plans will double the dimensions of the search space, which is impractical in our current framework;  
(2) We assume sticking finger contacts. Local planners for finger rolling/sliding can be considered for further dexterity; 
(3) The RRT is non-optimal, while the optimal RRT-star is not pratical as it is almost impossible to directly connect two nodes in our problem. Post-processing with trajectory optimization could be used to enhance the solution quality. Solutions from our planner can also be used as warm-start for CITO \cite{posa2014direct}; 
(4) We observed that some plans are fragile under uncertainty while others are very robust to execute even in an open-loop manner. It is important to have criteria over motion stability \cite{hou2020manipulation, johnson2016convergent} to increase the overall solution qualities;
(5) It is possible to adapt ideas and techniques of the CMGMP into other constrained sampling-based planning frameworks \cite{kingston2019exploring}.



\bibliographystyle{IEEEtran.bst}
\bibliography{IEEEabrv, references}
\end{document}